# Continuous Time Bayesian Networks

Uri Nodelman
Stanford University
nodelman@cs.stanford.edu

Christian R. Shelton
Stanford University
cshelton@cs.stanford.edu

Daphne Koller
Stanford University
koller@cs.stanford.edu

## Abstract

In this paper we present a language for finite state continuous time Bayesian networks (CTBNs), which describe structured stochastic processes that evolve over continuous time. The state of the system is decomposed into a set of local variables whose values change over time. The dynamics of the system are described by specifying the behavior of each local variable as a function of its parents in a directed (possibly cyclic) graph. The model specifies, at any given point in time, the distribution over two aspects: when a local variable changes its value and the next value it takes. These distributions are determined by the variable's current value and the current values of its parents in the graph. More formally, each variable is modelled as a finite state continuous time Markov process whose transition intensities are functions of its parents. We present a probabilistic semantics for the language in terms of the generative model a CTBN defines over sequences of events. We list types of queries one might ask of a CTBN, discuss the conceptual and computational difficulties associated with exact inference, and provide an algorithm for approximate inference which takes advantage of the structure within the process.

## 1 Introduction

Consider a medical situation where you have administered a drug to a patient and wish to know how long it will take for the drug to take effect. The answer to this question will likely depend on various factors, such as how recently the patient has eaten. We want to model the temporal process for the effect of the drug and how its dynamics depend on these other factors. As another example, we might want to predict the amount of time that a person remains unemployed, which can depend on the state of the economy, on their own financial situation, and more.

Although these questions touch on a wide variety of issues, they are all questions about distributions over time. Standard ways of approaching such questions — event history analysis (Blossfeld et al., 1988; Blossfeld & Rohwer, 1995; Anderson et al., 1993) and Markov process models (Duffie et al., 1996; Lando, 1998) — work well, but do not allow the specification of models with a large structured state space where some variables do not directly depend on others. For example, the distribution over how fast a drug takes effect might be mediated by how fast *it* reaches the bloodstream which may itself be affected by how recently the person has eaten.

Bayesian networks (Pearl, 1988) are a standard approach for modelling structured domains. With such a representation we can be explicit about the direct dependencies which are present and use the independencies to our advantage computationally. However, Bayesian networks are designed to reason about static processes, and cannot be used directly to answer the types of questions that concern us here.

Dynamic Bayesian networks (DBNs) (Dean & Kanazawa, 1989) are the standard extension of Bayesian networks to temporal processes. DBNs model a dynamic system by discretizing time and providing a Bayesian network fragment that represents the probabilistic transition of the state at time $t$ to the state at time $t + 1$. Thus, DBNs represent the state of the system at different points in time, but do not represent time explicitly. As a consequence, it is very difficult to query a DBN for a distribution over the time at which a particular event takes place. Moreover, since DBNs slice time into fixed increments, one must always propagate the joint distribution over the variables at the same rate. This requirement has several limitations. First, if our system is composed of processes that evolve at different time granularities, we must represent the entire system at the finest possible granularity. Second, if we obtain observations which are irregularly spaced in time, we must still represent the intervening time slices at which no evidence is obtained.

Hanks et al. (1995) present another discrete time approach to temporal reasoning related to DBNs which they extend with a rule-based formalism to model endogenous changes to variables which occur between exogenous events. They also include an extensive discussion of various approaches probabilistic temporal reasoning.

We provide the alternative framework of *continuous time Bayesian networks*. This framework explicitly represents temporal dynamics and allows us to query the network for the distribution over the time when particular events of interest occur. Given sequences of observations spaced irreg-ularly through time, we can propagate the joint distribution from observation to observation. Our approach is based on the framework of *homogeneous Markov processes*, but utilizes ideas from Bayesian networks to provide a graphical representation language for these systems. Endogenous changes are modelled by the state transitions of the process. The graphical representation allows compact models for processes involving a large number of co-evolving variables, and an effective approximate inference procedure similar to clique tree propagation.

## 2 Continuous Time

We begin with the necessary background on modelling with continuous time.

### 2.1 Homogeneous Markov Processes

Our approach is based on the framework of finite state continuous time Markov processes. Such processes are generally defined as matrices of transition *intensities* where the $(i, j)$ entry gives the intensity of transitioning from state $i$ to state $j$ and the entries along the main diagonal make each row sum to zero. Specifically, our framework will be based on *homogeneous Markov processes* — one in which the transition intensities do not depend on time.

Let $X$ be a local variable, one whose state changes continuously over time. Let the domain of $X$ be $Val(X) = \{x_1, x_2, \ldots, x_n\}$. We present a homogeneous Markov process $X(t)$ via its intensity matrix:

$$\mathbf{Q}_X = \begin{bmatrix} -q_1^x & q_{12}^x & \cdots & q_{1n}^x \\ q_{21}^x & -q_2^x & \cdots & q_{2n}^x \\ \vdots & \vdots & \ddots & \vdots \\ q_{n1}^x & q_{n2}^x & \cdots & -q_n^x \end{bmatrix},$$

where $q_i^x = -\sum_{j \neq i} q_{ij}^x$. Intuitively, the intensity $q_i^x$ gives the 'instantaneous probability' of leaving state $x_i$ and the intensity $q_{ij}^x$ gives the 'instantaneous probability' of transitioning from $x_i$ to $x_j$. More formally, as $\Delta t \to 0$,

$$\begin{aligned} \Pr\{X(t+\Delta t) = x_j \mid X(t) = x_i\} &\approx q_{ij}^x \Delta t, \text{for } i \neq j \\ \Pr\{X(t+\Delta t) = x_i \mid X(t) = x_i\} &\approx 1 - q_i^x \Delta t \end{aligned}.$$

Given the $\mathbf{Q}_X$ matrix we can describe the transient behavior of $X(t)$ as follows. If $X(0) = x_i$ then it stays in state $x_i$ for an amount of time exponentially distributed with parameter $q_i^x$. Thus, the probability density function $f$ and corresponding distribution function $F$ for $X(t)$ remaining equal to $x_i$ are given by

$$\begin{aligned} f(t) &= q_i^x \exp(-q_i^x t), & t \geq 0 \\ F(t) &= 1 - \exp(-q_i^x t), & t \geq 0 \end{aligned}.$$

The expected time of transitioning is $1/q_i^x$. Upon transitioning, $X$ shifts to state $x_j$ with probability $q_{ij}^x/q_i^x$.

**Example 2.1** *Assume that we want to model the behavior of the barometric pressure $B(t)$ discretized into three states ($b_1$ = falling, $b_2$ = steady, and $b_3$ = rising), we could write the intensity matrix as*

$$\mathbf{Q}_B = \begin{bmatrix} -.21 & .2 & .01 \\ .05 & -.1 & .05 \\ .01 & .2 & -.21 \end{bmatrix}.$$

*If we view units for time as hours, this means that if the pressure is falling, we expect that it will stop falling in a little less than 5 hours (1/.21 hours). It will then transition to being steady with probability .2/.21 and to falling with probability .01/.21.*

We can consider the transitions made between two consecutive different states, ignoring the time spent at each state. Specifically, we can define the embedded Markov chain $E$ which is formed by ignoring the amount of time $X$ spends in its states and noting only the sequence of transitions it makes from state to state. We can write out the $n \times n$ transition probability matrix $P_E$ for this chain, by putting zeros along the main diagonal and $q_{ij}^x/q_i^x$ in the $(i, j)$ entry. We can also consider the distribution over the amount of time $X$ spends in a state before leaving again, ignoring the particular transitions $X$ makes. We can write out the $n \times n$ state duration matrix $M$ (which is often called the *completion rate matrix* or *holding rate matrix*), by putting the $q_i$ values along the main diagonal and zeros everywhere else. It is easy to see that we can describe the original intensity matrix in terms of these two matrices:

$$\mathbf{Q} = M(P_E - I).$$

**Example 2.2** *For our barometric pressure process B,*

$$\mathbf{Q}_B = \begin{bmatrix} .21 & 0 & 0 \\ 0 & .1 & 0 \\ 0 & 0 & .21 \end{bmatrix} \left( \begin{bmatrix} 0 & \frac{20}{21} & \frac{1}{21} \\ \frac{1}{2} & 0 & \frac{1}{2} \\ \frac{20}{21} & \frac{1}{21} & 0 \end{bmatrix} - I \right).$$

### 2.2 Subsystems

It is often useful to consider *subsystems* of a Markov process. A subsystem, $S$, describes the behavior of the process over a subset of the full state space — i.e., $Val(S) \subset Val(X)$. In such cases we can form the intensity matrix of the subsystem, $U_S$, by using only those entries from $\mathbf{Q}_X$ that correspond to states in $S$.

**Example 2.3** *If we want the subsystem of the barometric pressure process, B, corresponding to the pressure being steady or rising ($S = \{b_2, b_3\}$), we get*

$$U_S = \begin{bmatrix} -.1 & .05 \\ .2 & -.21 \end{bmatrix}.$$

Note that, for a subsystem, the sums of entries along a row are not, in general, zeros. This is because a subsystem is not a closed system — i.e, from each state, there can be a positive probability of entering states not in $S$ and thus not represented in the transition matrix for the subsystem.

Once we have formed a subsystem $S$ of $X$, we can also talk about the complement subsystem $\bar{S}$, which is a subsystem over the other states — i.e., $Val(\bar{S}) = Val(X) - Val(S)$. In general, when examining the behavior of a subsystem, we consider the *entrance* and *exit* distributions for the subsystem. An *entrance distribution* is a distribution over the states of $S$, where the probability of a state $s$ is the probability that $s$ is the state to which we first transition when entering the subsystem $S$. An *exit distribution* describes the first state not in $Val(S)$ to which we transition when we leave the subsystem.

### 2.3 Queries over Markov processes

If we have an intensity matrix, $\mathbf{Q}_X$, for a homogeneous Markov process $X(t)$ and an initial distribution over the value of $X$ at time 0, $P_X^0$, there are many questions about the process which we can answer.

The conditional distribution over the value of $X$ at time $t$ given the value at an earlier time $s$ is

$$\Pr\{X(t) \mid X(s)\} = \exp(\mathbf{Q}_X(t-s)), \text{ for } s < t .$$

Thus, the distribution over the value of $X(t)$ is given by

$$P_X(t) = P_X^0 \exp(\mathbf{Q}_X t) .$$

As $t$ grows, $P_X(t)$ approaches the stationary distribution $\pi$ for $X$ which can be computed by an eigenvalue analysis. Additionally, we can form the joint distribution over any two time points using the above two formulas:

$$P_X(s,t) = P_X(s) \exp(\mathbf{Q}_X(t-s)) .$$

Suppose we are interested in some subsystem $S$ of $X$. Given an entrance distribution $P_S^0$ into $S$, we can calculate the distribution over the amount of time that we remain within the subsystem. This distribution function is called a *phase distribution* (Neuts 1975; 1981), and is given by

$$F(t) = 1 - P_S^0 \exp(U_S t) e.$$

where $U_S$ is (as above) the subsystem intensity matrix and $e$ is the unit vector. The expected time to remain within the subsystem is given by $P^0(-U_S)^{-1}e$.

**Example 2.4** *In our barometric pressure example, if we have a uniform entrance distribution for the subsystem in Example 2.3, then the distribution in time over when the pressure begins to fall is given by*

$$F(t) = 1 - [\ .5 \quad .5\ ] \exp\left(\begin{bmatrix} -.1 & .05 \\ .2 & -.21 \end{bmatrix} t\right) e$$
$$\approx 1 - 0.3466(-1.1025^t) - 0.6534(-0.1975^t) .$$

Finally, given an entrance distribution, $P_S^0$, to a subsystem $S$ of $X$, we can calculate the exit distribution. To do so, we construct a new process $X'$ by setting all intensities to zero within rows corresponding to states not in $S$. This transformation, in effect, makes every state which is not in the subsystem an absorbing state. (Once the system has entered an absorbing state, it can never leave that state.) If we use our entrance distribution over the states of $S$ for our initial distribution to $X'$ (setting the probability of starting in other states to zero), we can see that the exit distribution is given by the stationary distribution of $X'$. This is because the only way that we can enter the newly constructed absorbing states is by leaving $S$ and so the probability with which we end up in an absorbing state is the probability that we entered that state by exiting the subsystem.

## 3 Continuous Time Bayesian Nets

Our goal in this paper is to model Markov processes over systems whose momentary state is defined as an assignment to some (possibly large) set of variables $X$. In principle, we can simply explicitly enumerate the state space $Val(X)$, and write down the intensity matrix which specifies the transition intensity between every pair of these states. However, as in Bayesian networks, the size of the state space grows exponentially with the number of variables, rendering this type of representation infeasible for all but the smallest spaces.

In this section, we provide a more compact factored representation of Markov processes. We define a *continuous time Bayesian network* — a graphical model whose nodes are variables whose state evolves continuously over time, and where the evolution of each variable depends on the state of its parents in the graph.

### 3.1 Conditional Markov Processes

In order to compose Markov processes in a larger network, we need to introduce the notion of a *conditional Markov process*. This is a type of inhomogeneous Markov process where the intensities vary with time, but not as a direct function of time. Rather, the intensities are a function of the current values of a set of other variables, which also evolve as Markov processes. We note that a similar model was used by Lando (1998), but the conditioning variables were not viewed as Markov processes, nor were they built into any larger structured model, as in our framework.

Let $Y$ be a variable whose domain is $Val(Y) = \{y_1, y_2, \ldots, y_m\}$. Assume that $Y$ evolves as a Markov process $Y(t)$ whose dynamics are conditioned on a set $V$ of variables, each of which also can also evolve over time. Then we have a *conditional intensity matrix* (CIM) which can be written

$$\mathbf{Q}_{Y|V} = \begin{bmatrix} -q_1^y(V) & q_{12}^y(V) & \cdots & q_{1m}^y(V) \\ q_{21}^y(V) & -q_2^y(V) & \cdots & q_{2m}^y(V) \\ \vdots & \vdots & \ddots & \vdots \\ q_{m1}^y(V) & q_{m2}^y(V) & \cdots & -q_m^y(V) \end{bmatrix} .$$

Equivalently, we can view a CIM as set of intensity matrices, one for each instantiation of values $v$ to the variables $V$. The set of variables $V$ are called the *parents* of $Y$, and denoted $Par(Y)$. Note that, if the parent set $Par(Y)$ is empty, then the CIM is simply a standard intensity matrix.

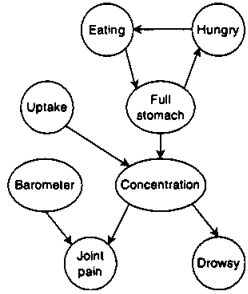

Figure 1: Drug effect network

**Example 3.1** *Consider a variable $E(t)$ which models whether or not a person is eating ($e_1$ = not eating, $e_2$ = eating) and is conditional on a variable $H(t)$ which models whether or not a person is hungry ($h_1$ = not hungry, $h_2$ = hungry). Then we can specify the CIM for $E(t)$ as*

$$\mathbf{Q}_{E|h_1} = \begin{bmatrix} -.01 & .01 \\ 10 & -10 \end{bmatrix} \quad \mathbf{Q}_{E|h_2} = \begin{bmatrix} -2 & 2 \\ .01 & -.01 \end{bmatrix}.$$

*Given this model, we expect a person who is hungry and not eating to begin eating in half an hour (1/2 hour). We expect a person who is not hungry and is eating to stop eating in 6 minutes (1/10 hour).*

### 3.2 The CTBN Model

Conditional intensity matrices provide us a way of modelling the local dependence of one variable on a set of others. By putting these local models together, we can define a single joint structured model. As is the case with dynamic Bayesian networks, there are two central components to define: the initial distribution and the dynamics with which the system evolves through time.

**Definition 3.2** *Let $X$ be a set of local variables $X_1, \ldots, X_n$. Each $X_i$ has a finite domain of values $\mathrm{Val}(X_i)$. A continuous time Bayesian network $\mathcal{N}$ over $X$ consists of two components: The first is an initial distribution $P_X^0$, specified as a Bayesian network $\mathcal{B}$ over $X$. The second is a continuous transition model, specified as*

- *A directed (possibly cyclic) graph $G$ whose nodes are $X_1, \ldots, X_n$; $\mathrm{Par}(X_i)$ denotes the parents of $X_i$ in $G$.*
- *A conditional intensity matrix, $\mathbf{Q}_{X_i|\mathrm{Par}(X_i)}$, for each variable $X_i \in X$.* ∎

Unlike traditional Bayesian networks, there is no problem with cycles in the graph $G$. An arc $X \to Y$ in the graph implies that the dynamics of $Y$'s evolution in time depends on the value of $X$. There is no reason why the dynamics of $X$'s evolution cannot simultaneously depend on the value of $Y$. This dependency is analogous to a DBN model where we have arcs $X^t \to Y^{t+1}$ and $Y^t \to X^{t+1}$.

**Example 3.3** *Figure 1 shows the graph structure for a CTBN modelling our drug effect example. There are nodes for the uptake of the drug and for the resulting concentration of the drug in the bloodstream. The concentration is affected by the how full the patient's stomach is. The drug is supposed to alleviate joint pain, which may be aggravated by falling pressure. The drug may also cause drowsiness. The model contains a cycle, indicating that whether a person is hungry depends on how full their stomach is, which depends on whether or not they are eating.*

### 3.3 Amalgamation

In order to define the semantics of a CTBN, we must show how to view the entire system as a single process. To do this, we introduce a "multiplication" operation called *amalgamation* on CIMs. This operation combines two CIMs to produce a single, larger CIM.

Amalgamation takes two conditional intensity matrices $\mathbf{Q}_{S_1|C_1}$ and $\mathbf{Q}_{S_2|C_2}$ and forms from them a new product CIM, $\mathbf{Q}_{S|C} = \mathbf{Q}_{S_1|C_1} * \mathbf{Q}_{S_2|C_2}$ where $S = S_1 \cup S_2$ and $C = (C_1 \cup C_2) - S$. The new CIM contains the intensities for the variables in $S$ conditioned on those of $C$. A basic assumption is that, as time is continuous, variables cannot transition at the same instant. Thus, all intensities corresponding to two simultaneous changes are zero. If the changing variable is in $S_1$, we can look up the correct intensity from the factor $\mathbf{Q}_{S_1|C_1}$. Similarly, if it is in $S_2$, we can look up the intensity from the factor $\mathbf{Q}_{S_2|C_2}$. Intensities along the main diagonal are computed at the end to make the rows sum to zero for each instantiation to $C$.

**Example 3.4** *Assume we have a CTBN with graph $W \leftrightarrows Z$ and CIMs*

$$\mathbf{Q}_{W|z_1} = \begin{bmatrix} -1 & 1 \\ 2 & -2 \end{bmatrix} \quad \mathbf{Q}_{W|z_2} = \begin{bmatrix} -3 & 3 \\ 4 & -4 \end{bmatrix}$$

$$\mathbf{Q}_{Z|w_1} = \begin{bmatrix} -5 & 5 \\ 6 & -6 \end{bmatrix} \quad \mathbf{Q}_{Z|w_2} = \begin{bmatrix} -7 & 7 \\ 8 & -8 \end{bmatrix}.$$

*Let us consider the joint transition intensity of these two processes. As discussed, intensities such as between $(z_1, w_1)$ and $(z_2, w_2)$ are zero. Now, consider a transition from $(z_1, w_1)$ to $(z_1, w_2)$. In this case, we simply use the appropriate transition intensity from the matrix $\mathbf{Q}_{W|z_1}$, i.e., 1. Assuming that the states in the joint space are ordered as $(z_1, w_1), (z_1, w_2), (z_2, w_1), (z_2, w_2)$, this would be the value of the row 1, column 2 entry of the joint intensity matrix. As another example, the value of the row 4, column 2 entry would be taken from $\mathbf{Q}_{Z|w_2}$. The entries on the diagonal are determined at the end, so as to make each row sum to 0. The joint intensity matrix is, therefore,*

$$\mathbf{Q}_{WZ} = \mathbf{Q}_{W|Z} * \mathbf{Q}_{Z|W} = \begin{bmatrix} -6 & 1 & 5 & 0 \\ 2 & -9 & 0 & 7 \\ 6 & 0 & -9 & 3 \\ 0 & 8 & 4 & -12 \end{bmatrix}.$$

To provide a formal definition, we need to introduce some notation. Let $\mathbf{Q}_{S|C}(s_i \to s_j \mid c_k)$ be the intensity specified in $\mathbf{Q}_{S|C}$ for the variables in $S$ changing from state $s_i$ to state $s_j$ conditioned on the variables of $C$ having value $c_k$.

We denote the set of variables whose values are different between the instantiations $s_i$ and $s_j$ as $\delta(i,j)$. We define $s[S_\ell]$ to be the projection of the instantiation $s$ onto the set of variables $S_\ell$. Finally, we use $(s_i, c_k)$ to denote the joint instantiation over $S, C$ consistent with $s_i, c_k$.

$$Q_{S|C}(s_i \to s_j \mid c_k)$$
$$= \begin{cases} Q_{S_1|C_1}(s_i[S_1] \to s_j[S_1] \mid (s_i, c_k)[C_1]) \\ \quad \text{if } |\delta(i,j)| = 1 \text{ and } \delta(i,j) \subseteq S_1 \\ Q_{S_2|C_2}(s_i[S_2] \to s_j[S_2] \mid (s_i, c_k)[C_2]) \\ \quad \text{if } |\delta(i,j)| = 1 \text{ and } \delta(i,j) \subseteq S_2 \\ -z_{i|k} \quad \text{if } i = j \\ 0 \quad \text{otherwise} \end{cases}$$

where $z_{i|k} = \sum_{j \neq i} Q_{S|C}(s_i \to s_j \mid c_k)$.

### 3.4 Semantics

Formally, let $(\Omega, \mathcal{F}, P)$ be our probability space, where the space $\Omega$ consists of a set of infinite trajectories over $\tau = [0, \infty)$, and $\mathcal{F}$ is an appropriate $\sigma$-algebra. (See Gihman and Skorohod (1971) for a formal definition.)

We define the semantics of a CTBN as a single homogeneous Markov process over the joint state space, using the amalgamation operation. In particular, the CTBN $\mathcal{N}$ is a factored representation of the homogeneous Markov process described by the *joint intensity matrix* defined as

$$\mathbf{Q}_\mathcal{N} = \prod_{X \in \mathbf{X}} \mathbf{Q}_{X|Par(X)}.$$

¿From the definition of amalgamation, we see the states of the joint intensity matrix are full instantiations to all of the variables $\mathbf{X}$ of $\mathcal{N}$. Moreover, a single variable $X \in \mathbf{X}$ transitions from $x_i$ to $x_j$ with intensity $q_{ij}^x(Par(X))$.

An alternative view of CTBNs is via a generative semantics. A CTBN can be seen as defining a generative model over sequences of *events*, where an *event* is a pair $\langle X \leftarrow x_j, T \rangle$, which denotes a transition of the variable $X$ to the value $x_j$ at time $T$. Given a CTBN $\mathcal{N}$, we can define the generative model as follows.

We initialize $\sigma$ to be an empty event sequence. We define a temporary event list $E$, which contains pairs of state transitions and intensities; the list $E$ is a data structure for candidate events, which is used to compute the next event in the event sequence. We also maintain the current time $T$ and the current state of the system $x(T)$. Initially, we have $T = 0$, and the system state $x(0)$ is initialized by sampling at random from the Bayesian network $\mathcal{B}$ which denotes the initial state distribution $P_X^0$.

We then repeat the following steps, where each step selects the next event to occur in the system, with the appropriate distribution.

For each variable $X$ that does not have an event in $E$:
    Let $x_i = x(t)[X]$
    Choose the transition $x_i \to x_j$ according to the
        probabilities $q_{ij}^x(Par(X))/q_i^x(Par(X))$
    Add $\langle X \leftarrow x_j, q_i^x(Par(X)) \rangle$ to $E$

Let $q_E$ be the sum of all the $q$ values for events in $E$
Choose the next event $\langle X \leftarrow x_j, q^x \rangle$ from $E$ with
    probability $q^x/q_E$
Choose the time $t_E$ for the next transition from an
    exponential distribution with parameter $q_E$.
Update $T \leftarrow T + t_E$ and $X \leftarrow x_j$
Add $\langle X \leftarrow x_j, T \rangle$ to $\sigma$
Remove from $E$ the transition for $X$ and for all
    variables $Y$ for which $X \in Par(Y)$.

**Definition 3.5** *Two Markov processes are said to be* stochastically equivalent *if they have the same state space and transition probabilities (Gihman & Skorohod, 1973).*

**Theorem 3.6** *The Markov process determined by the generative semantics is stochastically equivalent to the Markov process determined by the joint intensity matrix.*

As in a Bayesian network, the graph structure can be viewed in two different yet closely related ways. The first is as a data structure with which we can associate parameters to define a joint distribution. The second is as a qualitative description of the independence properties of the distribution. To understand this notion, note that there are two ways to think about a stochastic process $X$. For a fixed time $t \in \tau$, $X$ can be viewed as a random variable $X(t)$. For a fixed $\omega \in \Omega$, we can view $X$ as a function of time (over $\tau$) and $X(\omega)$ as a trajectory. The CTBN graph specifies a notion of independence over entire trajectories.

**Definition 3.7** *Two Markov processes $X$ and $Y$ are* independent *if, for any finite sets $T, T' \subset \tau$, the joint distribution over the variables $X(t), t \in T$ and the joint distribution over the variables $Y(t'), t' \in T'$ are independent (Gihman & Skorohod, 1971).*

**Definition 3.8** *We say that $Y$ is a* descendants *of $X$ in the (possibly cyclic) graph $G$ if and only if there is a directed path in $G$ from $X$ to $Y$. (Note that a variable can be its own descendants according to this definition.)*

**Theorem 3.9** *If $X$ is a local variable in a CTBN $\mathcal{N}$, then $X$ is independent of its non-descendants (in $G$) given trajectories over the set of variables in $Par(X)$.*

For example, in our drug effect network, the joint pain is independent of taking the drug given the moment by moment concentration of the drug in the bloodstream.

## 4 Reasoning in CTBNs

In this section, we describe some of the queries that we can address using this type of representation. We then discuss some of the computational difficulties that we encounter if we try doing this type of inference exactly.

### 4.1 Queries over a CTBN

In the previous section, we showed that we can view a CTBN as a compact representation of a joint intensity matrix for a homogeneous Markov process. Thus, at least in principle, we can use a CTBN to answer any query that we can answer using an explicit representation of a Markov process: We can form the joint intensity matrix and then answer queries just as we do for any homogeneous Markov process, as described above.

For example, in the drug effect network, we can set the initial distribution such that the drug was administered at $t = 0$ *hours*, compute the joint distribution over the state of the system at $t = 5$, and then marginalize it to obtain a distribution over joint pain at $t = 5$. Additionally, because we have the full joint distribution at this point in time, we can calculate for $t = 5$ the distribution over drowsiness given that the concentration of the drug is high.

Now, assume that we have a series of observations. We can compute the joint distribution over the system state for any point in time at or after the time of the last observation. We calculate the new joint distribution at the time of the first observation, condition on the observation, and use that as the initial distribution from which to compute the joint distribution at the next observation time. This process can be executed for an entire series of observations. For example, assume that our patient took the drug at $t = 0$, ate after an hour ($t = 1$) and felt drowsy three hours after eating ($t = 4$). We can compute the distribution over joint pain six hours after taking the drug ($t = 6$) by computing the joint distribution at time 1, conditioning that distribution on the observation of eating, and using that as an initial distribution with which to compute the joint distribution 3 hours later. After conditioning on the observation of drowsiness, we use the result as an initial distribution with which to calculate the joint distribution 2 hours after that. That joint distribution can be marginalized to give the distribution over joint pain given the sequence of evidence. The key is that, unlike in DBNs, we need only do one propagation for each observation time, even if the observations are irregularly spaced.

As noted in section 2.3 we can compute the joint distribution between any two points in time. By conditioning on evidence at the later time point, we can propagate evidence backwards in time. Even more interestingly, we can calculate the distribution over the first time a variable $X$ takes on a particular value $x$: $X$ taking the value $x$ is simply a subsystem of the joint intensity matrix, and we can compute the distribution over the entrance time into the subsystem. For example, we could set our initial distribution to one where the patient takes the drug and has joint pain. We could then directly compute the distribution over the time at which the joint pain goes away. Note that this type of query could also be computed for the time after some sequence of evidence.

### 4.2 Difficulties with Exact Inference

The obvious flaw in our discussion above is that our approach for answering these queries requires that we generate the full joint intensity matrix for the system as a whole, which is exponential in the number of variables. The graphical structure of the CTBN immediately suggests that we perform the inference in a decomposed way, as in Bayesian networks. Unfortunately, as we now show, the problems are significantly more complex in this setting.

Consider a simple chain $X \to Y \to Z$. It might appear that, at any point in time, $Z$ is independent of $X$ given $Y$. Unfortunately, this is not the case. Even though the transition intensity for $Z$ depends only on the value of $Y$ at any instance in time, as soon as we consider temporal evolution, their states become correlated. This problem is completely analogous to the entanglement problem in DBNs (Boyen & Koller, 1998), where all variables in the DBN typically become correlated over some number of time slices. The primary difference is that, in continuous time, even the smallest time increment $\Delta t$ results in the same level of entanglement as we would gain from an arbitrary number of time slices in a DBN.

In fact, as discussed in Section 3.4, the only conclusion we can make about a structure $X \to Y \to Z$ is that the $Z$ is independent of $X$ given the *full trajectory* of $Y$. As a consequence, we can fully reconstruct the distribution over trajectories of $Z$, ignoring $X$, if we are given the full distribution over trajectories for $Y$. Of course, a full distribution over continuous time processes is a fairly complex structure. One might hope that we can represent it compactly, e.g., using an intensity matrix. Unfortunately, even when the distribution over the joint $X, Y$ process is a homogeneous Markov process, its projection over $Y$ is not a homogeneous Markov process.

A second potential avenue is the fairly natural conjecture that we do not always need the full distribution over trajectories. Perhaps, if our goal is only to answer certain types of queries, we can make do with some summary over $Y$. Most obviously, suppose we want to compute the stationary distribution over $Z$. It seems reasonable to assume that $Z$'s stationary behavior might depend only on the stationary behavior of $Y$. After all, the transitions for $Z$ are governed by two matrices $\mathbf{Q}_{Z|y_1}$ and $\mathbf{Q}_{Z|y_2}$. As long as we know the stationary distribution for $Y$, we know which fraction of the time $Z$ uses each of its transition matrices. So, we should be able to compute the stationary distribution for $Z$ from this information. Unfortunately, this assumption turns out to be unfounded.

**Example 4.1** *Consider the following intensity matrices.*

$$\mathbf{Q}_Y = \begin{bmatrix} -1 & 1 \\ 2 & -2 \end{bmatrix} \quad \mathbf{Q}_{Y'} = \begin{bmatrix} -10 & 10 \\ 20 & -20 \end{bmatrix}$$

$$\mathbf{Q}_{Z|y_1} = \begin{bmatrix} -3 & 3 \\ 15 & -15 \end{bmatrix} \quad \mathbf{Q}_{Z|y_2} = \begin{bmatrix} -5 & 5 \\ 4 & -4 \end{bmatrix}$$

*Note that $Y$ and $Y'$ both have the same stationary distribution, $\begin{bmatrix} .75 & .25 \end{bmatrix}$. If we look at the CTBN with the graph $Y \to Z$ we get a stationary distribution for $Z$ of $\begin{bmatrix} .7150 & .2850 \end{bmatrix}$. But, if we look at the CTBN with graph $Y' \to Z$, we get a stationary distribution for $Z$ of $\begin{bmatrix} .7418 & .2582 \end{bmatrix}$.*

Thus, even the stationary behavior of $Z$ depends on the specific trajectory of $Y$ and not merely the fraction of time it spends in each of its states. We can gain intuition for this phenomenon by thinking about the intensity matrix as an infinitesimal transition matrix. To determine the behavior of $Z$, we can imagine that for each infinitesimal moment of time we multiply it to get to the next time instance.[1] At each instance, we check the value of $Y$ and select which matrix we multiply for that instant. The argument that we can restrict attention to the stationary distribution of $Y$ implicitly assumes that we care only about "how many" times we use each matrix. Unfortunately, matrix multiplication does not commute. If we are rapidly switching back and forth between different values of $Y$ we get a different product at the end than if we switch between the values more slowly. The product is different because the order in which we multiplied was different — even if the number of times we used one matrix or the other were same in both cases.

## 5 Approximate Inference

As we saw in the previous section, exact inference in CTBNs is probably intractable. In this section, we describe an approximate inference technique based on the clique tree inference algorithm. Essentially, the messages passed between cliques are distributions over entire trajectories, represented as homogeneous Markov processes. These messages are not the correct distributions, but they often provide a useful approximation.

For example, consider again our chain $X \to Y \to Z$. As we mentioned, to reason about $X$, we need to pass to $Z$ the entire distribution over $Y$'s trajectories. Unfortunately, the projection of the entire process onto $Y$ is not a homogeneous Markov process. However, we can build a process over $X, Y$, and *approximate* the distribution over $Y$'s trajectories as a homogeneous Markov process.

### 5.1 The Clique Tree Algorithm

Roughly speaking, the basic clique tree calibration step is almost identical to the propagation used in the Shafer and Shenoy (1990) algorithm, except that we use amalgamation as a substitute for products and approximate marginalization as a substitute for standard marginalization.

**Initialization of the Clique Tree** We begin by constructing the clique tree for the graph $G$. This procedure is the same as with ordinary Bayesian networks except that we must deal with cycles. We simply connect all parents of a node with undirected edges, and then make all the remaining edges undirected. If we have a cycle, it simply turns into a loop in the resulting undirected graph.

As usual, we associate each variable with a clique that contains it and all of its parents, and assign its CIM to that clique. Let $A_i \subseteq C_i$ be the set of variables associated with clique $C_i$. Let $N_i$ be the set of neighboringcliques for $C_i$ and let $S_{ij}$ be the set of variables in $C_i \cap C_j$. We also compute, for each clique $C_i$, the initial distribution $P_i(C_i)$. We can implement this operation using standard BN inference on the network $\mathcal{B}$. Finally, we calculate the initial intensity potential $f_i$ for $C_i$ as:

$$f_i = \prod_{X \in A_i} \mathbf{Q}_{X|Par(X)}$$

where our notion of product is amalgamation.

**Message Passing** The message passing process is used purely for initial calibration. Its basic goal is to compute, in each clique $i$, an approximate probability distribution over the trajectories of the variables $C_i$. This approximation is simply a homogeneous Markov process, and is represented as an initial distribution (computed in the initialization step) and a joint intensity matrix over $C_i$, computed in the calibration step (described below). At this point in the algorithm, no evidence is introduced.

To perform the calibration, cliques send messages to each other. A clique $C_i$ is ready to send message $\mu_{i \to j}$ to clique $C_j$ when it has received messages $\mu_{k \to i}$ from all the neighboring cliques $k \in N_i$ except possibly $j$. At that point, we compute and send the message by amalgamating the local intensity potential with the other messages and eliminating all variables except those shared with $C_j$. More formally:

$$\mu_{i \to j} = \mathrm{marg}^{P_i}_{(C_i - S_{ij})} \left( f_i * \left( \prod_{k \in N_i, k \neq j} \mu_{k \to i} \right) \right),$$

where $\mathrm{marg}^P_Y(\mathbf{Q}_{S|C})$ denotes the operation of taking a CIM $\mathbf{Q}_{S|C}$ and eliminating the variables in $Y$. As we discussed, we cannot compute an exact representation of a Markov process after eliminating some subset of the variables. Therefore, we will use an approximate marginalization operation, which we describe below.

Once clique $C_i$ has received all of its incoming messages, we can compute a local intensity matrix as

$$\mathbf{Q}_{C_i} = f_i * \prod_{k \in N_i} \mu_{j \to i}$$

**Answering queries** After the calibration process, each clique $i$ has a joint intensity matrix over the variables in $C_i$, which, together with the initial distribution, define a homogeneous Markov process. We can therefore compute the approximate behavior of any of the variables in the clique, and answer any of the types of queries described in Section 4.1, for variables within the same clique.

Incorporating evidence is slightly more subtle. Assume that we want to introduce evidence over some variable $X$ at time $t_e$. We can reason over each Markov process separately to compute a standard joint distribution $P_{t_e}(C_i)$ over each clique $C_i$ at the time point $t_e$. However, as our approximation is different in different cliques, the distributions over different cliques will not be calibrated: The same variable at different cliques will typically have a different marginal distribution. In order to calibrate the clique tree to define a single coherent joint distribution, we must decide on a root $C_r$ for the tree, and do a standard downward pass from $C_r$ to calibrate all of the cliques to $C_r$. Once that is done, we have a coherent joint distribution, into which we can insert evidence, and which we can query for the probability of any variable of interest.

---

[1] The product integral can be used to make this argument mathematically precise (Gill & Johansen, 1990).

If we have a sequence of observations at times $t_1, \ldots, t_n$, we use the process described above to propagate the distribution to time $t_1$, we condition on the evidence, and then we use the new clique distributions at time $t_1$ as initial distributions from which we propagate to $t_2$. This process is repeated until we have propagated to time $t_n$ and incorporated the evidence, after which we are ready to answer queries that refer to time(s) after the last evidence point.

For evidence after the query time, we propagate the evidence from the final evidence point backward, iterating to the query time, constructing the probability of the later evidence given the query. Multiplying this by the forward propagation from the previous paragraph yields the probability of the query conditioned on all of the evidence.

### 5.2 Marginalization

The core of our algorithm is an "approximate marginalization" operation on intensity matrices which removes a variable ($X$ in our example) from an intensity matrix and approximates the resulting distribution over the other variables using a simpler intensity matrix. More formally, the marginalization operation takes a CIM $\mathbf{Q}_{S|C}$, a set of variables $Y \subset S$, and an initial distribution $P$ over the variables of $S$. It returns a reduced CIM of the form $\mathbf{Q}_{S'|C} = \mathrm{marg}_Y^P(\mathbf{Q}_{S|C})$, where $S' = S - Y$.

#### 5.2.1 The Linearization Method

Ideally, the transition probabilities derived from the marginalized matrix $\mathbf{Q}_{S'|C}$ would be equal to the actual transition probabilities derived from the original matrix $\mathbf{Q}_{S|C}$. Let $s' \oplus y$ be the full instantiation to $S$ of $s'$ to $S'$ and $y$ to $Y$. Consider the transition from $s_1 = s'_1 \oplus y_1$ to $s_2 = s'_2 \oplus y_2$ over an interval of length $\Delta t$. We would like our marginalized process to obey

$$P(s'_2|s'_1,c) = \sum_{y_1,y_2} P(s'_2 \oplus y_1|s'_1 \oplus y_2, c)P(y_1|s'_1,c)$$

for all $\Delta t$, $s_1$, $s_2$, and $c$. As discussed above, this is generally not possible.

Our linearization approach is based on two approximations. First, we assume that the value of the variables in $Y$ do not change over time, so that we can use the values of $y$ at the beginning of the interval:

$$P(s'_2|s'_1,c) \approx \sum_y P(s'_2 \oplus y|s'_1 \oplus y, c)P^0(y|s'_1,c),$$

where $P^0$ is the distribution at the beginning of the interval.

Second, we use a linear approximation to the matrix exponential:

$$\exp(\mathbf{Q}\Delta t) \approx I + \mathbf{Q}\Delta t .$$

The resulting approximation is

$$\mathbf{Q}_{S'|C}(s'_1 \to s'_2 \mid c)$$
$$\approx \sum_y \mathbf{Q}_{S|C}(s'_1 \oplus y \to s'_2 \oplus y \mid c)P^0(y \mid s'_1,c)$$

We call this expression the *linear approximation of the marginal*.

**Example 5.1** *Consider the CIMs from Example 4.1; amalgamated into a single system, we get:*

$$\mathbf{Q}_{YZ} = \begin{bmatrix} -4 & 1 & 3 & 0 \\ 2 & -7 & 0 & 5 \\ 15 & 0 & -16 & 1 \\ 0 & 4 & 2 & -6 \end{bmatrix} .$$

*If $P_Y^0 = \begin{bmatrix} .3 & .7 \end{bmatrix}$, $P_{Z|y_1}^0 = \begin{bmatrix} .7 & .3 \end{bmatrix}$, and $P_{Z|y_2}^0 = \begin{bmatrix} .3 & .7 \end{bmatrix}$, then the linear approximation is*

$$\mathrm{marg}_Y^{P^0}(\mathbf{Q}_{YZ}) = \begin{bmatrix} -4 & 4 \\ 5.6101 & -5.6101 \end{bmatrix} .$$

#### 5.2.2 The Subsystem Method

Unfortunately, unless we plan to do our inference with a significant amount of time slicing, the assumptions underlying the above method do not hold. In particular, if we want our approximation to work better over a longer time interval, we need to account for the fact that the variables we are eliminating can change over time. To do this, we will sacrifice some accuracy over short time intervals — which can be seen in Section 6.

To compute the subsystem approximation of the marginal, we first consider each assignment of values $s'$ to $S'$. We take all states $s$ that are consistent with $s'$ (which correspond to the different assignments to $Y$), and collapse them into a single state (or row of the intensity matrix). To understand the approximation, we recall that our approximate intensity matrix $\mathbf{Q}$ can be written as $M(P_E - I)$ where $M$ is the matrix describing the distribution over how much time we spend in a particular state and $P_E$ is the transition matrix for the embedded Markov chain, which determines the distribution over the value to which we transition. We approximate these two distributions for each subsystem and then form the new reduced matrix by multiplying.

Our reduced subsystem corresponding to $X'$ has only a single state, so its entry in the reduced holding matrix $M$ will be only a single parameter value corresponding to the momentary probability of simply staying in the same state. In our original intensity matrix, this parameter corresponds to the probability of staying within the subsystem. Our approximation chooses this parameter so as to preserve the expected time that we stay within the subsystem.

The transition matrix $P_E$ represents the transition matrix for the chain. Again, as our collapsed system has only a single state, we are only concerned with parameterizing the intensities of transitioning from the new state to states outside the subsystem. These probabilities are precisely those that characterize the exit distribution from the subsystem.

Before providing the exact formulas for these two computations, we recall that both the holding time for a subsystem and its exit distribution depend on the distribution with which we enter the subsystem. Over time, we can enter the subsystem multiple times, and the entrance distribution differs each time. For the approximation, however, we must pick a single entrance distribution. There are several choices that we can consider for an approximate entrance distribution. One simple choice is to take the initial distribution and use the portion which corresponds to being in the subsystem at the initial time point. We could also use the appropriate portion of the distribution at a later time, $t^*$.

Given an entrance distribution $P^0$, we can now compute both the holding time and the exit distribution. Recall that the distribution function over the holding time within a subsystem is given by:

$$F(t) = 1 - P^0 \exp(U_S t) e$$

In order to preserve the expected holding time, we must set our holding intensity value to be:

$$P^0(-U_S)^{-1} e.$$

The approximation for the $P_E$ matrix is a single row vector corresponding to the exit distribution from the subsystem, using $P^0$ as our entrance distribution.

**Example 5.2** *Consider again the system from Example 5.1. The subsystem approximation for $t^* = 0$ is*

$$marg_Y^{P^0}(\mathbf{Q}_{YZ}) = \begin{bmatrix} -3.7143 & 3.7143 \\ 5.7698 & -5.7698 \end{bmatrix}.$$

$1/3.7143 = 0.2692$ *is the expected holding time in the subsystem*

$$U_S = \begin{bmatrix} -4 & 1 \\ 2 & -7 \end{bmatrix},$$

*which corresponds to* $Z = z_1$. *In particular, $P^0_{Y|z_1}(-U_S)^{-1} e = 0.2692$ where we have calculated $P^0_{Y|z_1} = \begin{bmatrix} .5 & .5 \end{bmatrix}$.*

## 6 Experimental Results

For our experiments, we used the example network described in Figure 1. We implemented both exact inference and our approximation algorithm and compared the results. In our scenario at $t = 0$, the person modelled by the system experiences joint pain due to falling barometric pressure and takes the drug to alleviate the pain, is not eating, has an empty stomach, is not hungry, and is not drowsy. The drug is uptaking and the current concentration is 0.

We consider two scenarios. For both, Figure 2 shows the resulting distribution over joint pain as a function of time and the KL-divergence between the true joint distribution and the estimated joint distribution. In the first scenario (top row of plots), no evidence is observed. In the second scenario (bottom row of plots), we observe at $t = 1$ that the person is not hungry and at $t = 3$, that he is drowsy.

In both cases, (a) compares the exact distribution with the approximate distribution for both marginalization methods (linear and subsystem) and differing values of $t^*$ for the subsystem approximation. In both cases, we used a single approximation for the entire trajectory between evidence points. By contrast, (b) compares the same approximations when the dynamics are repeatedly recalculated at regular intervals by using the estimated distribution at the end of one interval as the new entrance distribution for the approximate dynamics of the next interval. The graph shows this recomputation at both 1hr and 6min intervals. The final graph (c) shows the average KL-divergence between the true joint distribution and the approximate joint distributions, averaged over 60 time points between $t = 0$ and $t = 6$, as the number of (evenly spaced) recalculation points grows for both marginalization methods.

As we can see in all of the results, the subsystem approximation performs better for longer time segments. However, when the time-slicing becomes too fine, the errors from this method grow. By comparison, the linear approximation performs very poorly as an approximation for long intervals, but its accuracy improves as the time granularity becomes finer. These results are in accordance with the intuitions used to build each approximation. The linear approximation makes two assumptions that only hold over short time intervals: eliminated variables do not change during the interval and the exponentiation of a matrix can be linearly approximated. By comparison, the subsystem approximation allows for multiple variables to change over the interval of approximation but, in doing so, gives up accuracy for small time scales.

## 7 Discussion

We have described a new modelling language for structured stochastic processes which evolve in continuous time. Because time is explicitly represented in the model, we can reason with it directly, and even answer queries which ask for a distribution over time. Moreover, the continuous time model enables us to deal with sequences of evidence by propagating the distribution over the values from the time of one observation to the next — even when the evidence is not evenly spaced.

To compare the CTBN and DBN frameworks, suppose we start with a non-trivial CTBN. For any finite amount of time, probabilistic influence can flow between any variables connected by a path in the CTBN graph. Thus, if we want to construct an "equivalent" DBN, the 2-TBN must be fully connected regardless of the $\Delta t$ we choose for each time slice. We can construct a DBN that approximates the CTBN by picking a subset of the connections (*e.g.* those which have the strongest influence). Yet, we still have the standard problem of exponential blowup in performing inference over time. So we would be led to perform approximate DBN inference in an approximate DBN. While this could form the basis of an approximation algorithm for CTBNs, we chose to work directly with continuous time, making a direct approximation.

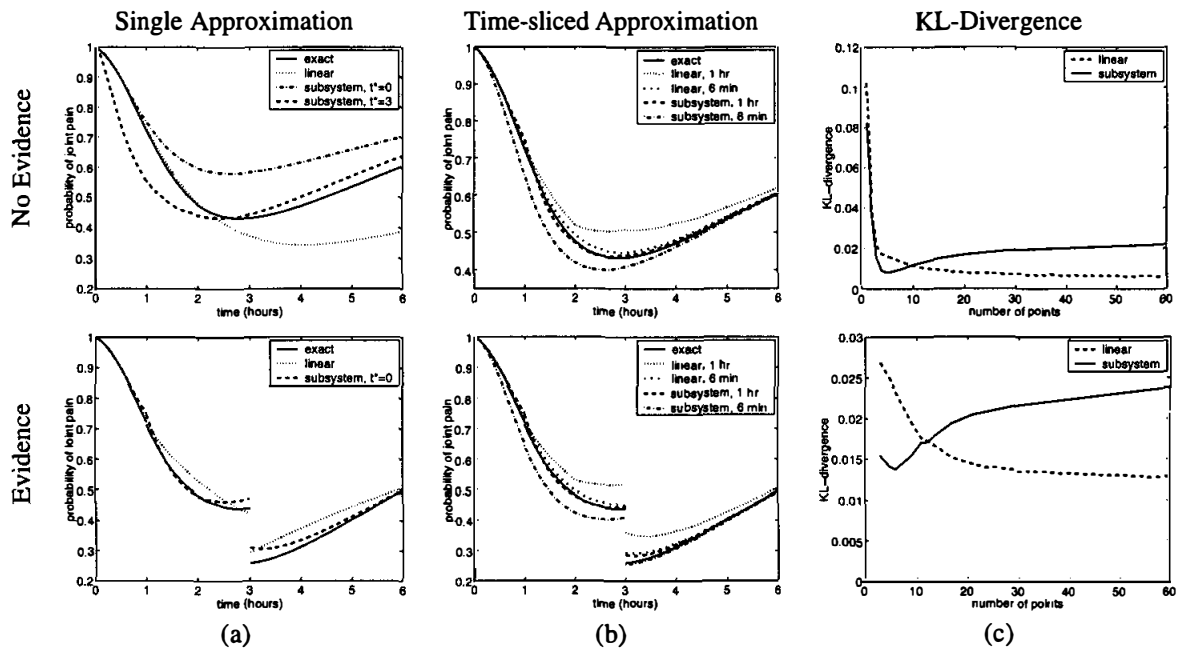

Figure 2: For the case of no evidence (top) and evidence (bottom): (a) The distribution for joint pain for exact and approximate inference without recalculation, (b) The distribution for joint pain for exact and approximate inference with recalculation of the dynamics every hour and every 6 minutes, and (c) The KL-divergence between the true distribution and the estimated distribution over all variables.

There are still several types of queries which we cannot yet answer. We cannot deal with situations where the evidence has the form "$X$ stayed at the value $x$ for the entire period between $t_1$ and $t_2$." Nor can we answer queries when we are interested in the distribution over $Y$ at the time when $X$ first transitions to $x_1$.

There are also many other important open questions. These include a theoretical analysis of the computational properties of these models and a more systematic theoretical and empirical analysis of the nature of our approximation algorithm, leading perhaps to a more informed method for choosing the entrance distribution for the marginalization operation. As an alternative approach, the generative semantics for CTBNs provides a basis for a stochastic sampling based method to approximate, which we would like to extend to situations with evidence using techniques such as importance sampling or MCMC. Even more globally, we would like to pursue parameter learning and structure learning of CTBNs from data.

**Acknowledgments** We would like to thank Carlos Guestrin and Balaji Prabhakar for their useful comments and suggestions. This work was funded by ONR contract N00014-00-1-0637 under the MURI program "Decision Making under Uncertainty."

## References

Anderson, P. K., Borgan, Ø., Gill, R. D., & Keiding, N. (1993). *Statistical models based on counting processes*. Springer-Verlag.

Blossfeld, H.-P., Hamerle, A., & Mayer, K. U. (1988). *Event history analysis*. Lawrence Erlbaum Associates.

Blossfeld, H.-P., & Rohwer, G. (1995). *Techniques of event history modeling*. Lawrence Erlbaum Associates.

Boyen, X., & Koller, D. (1998). Tractable inference for complex stochastic processes. *Proc. 14th UAI* (pp. 33–42).

Dean, T., & Kanazawa, K. (1989). A model for reasoning about persistence and causation. *Comp. Intelligence*, 5, 142–150.

Duffie, D., Schroder, M., & Skiadas, C. (1996). Recursive valuation of defaultable securities and the timing of resolution of uncertainty. *The Annals of Applied Probability*, 6, 1075–1090.

Gihman, I. I., & Skorohod, A. V. (1971). *The theory of stochastic processes I*. Springer-Verlag.

Gihman, I. I., & Skorohod, A. V. (1973). *The theory of stochastic processes II*. Springer-Verlag.

Gill, R. D., & Johansen, S. (1990). A survey of product-integration with a view towards applications in survival analysis. *The Annals of Statistics*, 18, 1501–1555.

Hanks, S., Madigan, D., & Gavrin, J. (1995). Probabilistic temporal reasoning with endogenous change. *Proc. 11th UAI*.

Lando, D. (1998). On Cox processes and credit risky securities. *Review of Derivatives Research*, 2, 99–120.

Neuts, M. F. (1975). Probability distributions of phase type. *Liber Amicorum Prof. Emeritus H. Florin* (pp. 173–206).

Neuts, M. F. (1981). *Matrix-geometric solutions in stochastic models — an algorithmic approach*. JHU Press.

Pearl, J. (1988). *Probabilistic reasoning in intelligent systems*. Morgan Kauffman.

Shafer, G., & Shenoy, P. (1990). Probability propagation. *Annals of Mathematics and Artificial Intelligence*, 2, 327–352.